\documentclass[runningheads]{llncs}

\usepackage{eccv}

\usepackage{eccvabbrv}

\usepackage{graphicx}
\usepackage{booktabs}

\usepackage[accsupp]{axessibility}  

\usepackage{hyperref}

\usepackage{orcidlink}

\usepackage{booktabs}
\usepackage{graphicx} %
\usepackage[dvipsnames]{xcolor}
\usepackage{algorithm}
\usepackage{mathtools}
\usepackage{svg}
\usepackage{algpseudocode}
\usepackage[ruled,vlined,algo2e]{algorithm2e}
\usepackage{marvosym}
\usepackage{multirow}
\usepackage{color}
\usepackage{colortbl}
\usepackage[font=small,skip=0pt]{caption}
\usepackage{float}
\usepackage{fontawesome}
\usepackage{enumitem}
\usepackage{makecell}
\usepackage{microtype}
\usepackage{wrapfig}
\newcommand{\coolname}{\textit{GhostPoint}}

\newcommand{\PAR}[1]{\vskip2pt \noindent{\bf #1}}
\newcolumntype{?}{!{\vrule width 1pt}}

\begin{document}

\title{GhostPoint: Self-Supervised Representation Learning by Hallucinating Occluded LiDAR Structure} 


\titlerunning{\textbf{GhostPoint}}

\makeatletter
\newcommand{\printfnsymbol}[1]{%
  \textsuperscript{\@fnsymbol{#1}}%
}
\makeatother

\author{Mohamed Abdelsamad\inst{1,2}\thanks{Equal contribution.} \and
Bin Yang\inst{1,3}\printfnsymbol{1} \and
Michael Ulrich\inst{1} \and
Miao Zhang\inst{1} \and
Yakov Miron\inst{1} \and
Alexandru Paul Condurache\inst{1,3} \and
Abhinav Valada\inst{2}}

\authorrunning{M. Abdelsamad et al.}

\institute{Bosch Center for AI \and University of Freiburg \and University of Luebeck}

\maketitle

\begin{abstract}
3D object detection from LiDAR point clouds is a core problem in autonomous driving. Recent advances in self-supervised learning (SSL) enable scalable pretraining and transfers well to per-point tasks such as semantic and panoptic segmentation, but transfer to 3D detection remains weaker.
We analyze recent SSL methods and find that most objectives are defined only on measured LiDAR returns from visible surfaces, leaving occluded and unobserved regions unconstrained. This visible-surface bias can be sufficient for point-wise prediction, but 3D detection requires robustness to missing structure. To address this gap, we propose \coolname{}, an SSL framework that hallucinates latent features in local neighborhoods around discovered instances, generated via a novel instance voxel dilation. In \coolname{}, an encoder processes observed returns, and an additional predictor infers neighborhood representations from observed context. In addition to standard encoder-level supervision, we introduce a predictor-level supervision scheme on sampled voxels from generated neighborhoods. Specifically, observed (visible/masked) voxels match teacher-encoder targets, while unobserved voxels match teacher-predictor hallucinations. This design encourages the learned representation to explicitly model structure beyond observed returns. Extensive evaluations on nuScenes and Waymo demonstrate that our method achieves state-of-the-art performance, consistently improving downstream 3D detection, especially under sparse scans and limited labels.
\end{abstract}

\section{Introduction}
\label{sec:intro}

\begin{figure}[t]
\centering
    \includegraphics[width=\columnwidth]{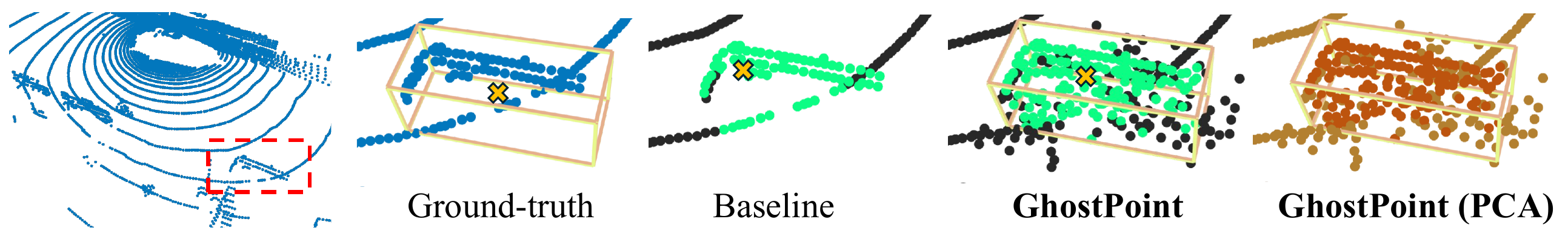}
    \vspace{0.1em}
\caption{\coolname{} learns beyond visible LiDAR surfaces. Raw scans observe only sparse object returns, causing discovered pseudo-instance centers to be biased toward visible surfaces. From these pseudo instances, \coolname{} samples adjacent unobserved regions and hallucinates their latent representations, producing features whose centers better align with the ground-truth object centers, marked by yellow crosses, while recovering object-level structure as illustrated by the PCA visualization.}
\vspace{-1mm}
\label{fig:empirical_gap}
\end{figure}

A robust 3D object detector is central to autonomous driving as it provides explicit object extents, position, and orientation for safety-critical behaviors such as collision avoidance, motion planning, and tracking in dynamic traffic scenes~\cite{buchner20223d,gosala2026sparse3dtrack,kappeler2026leveraging}. Its relevance remains high even in the era of end-to-end autonomous driving, where explicit 3D perception still offers strong geometric priors and safety interpretability~\cite{hu2023uniad,zhang2025bridgead}. Accordingly, recent LiDAR-centric detection research continues to improve robustness and label efficiency in realistic driving regimes~\cite{zhan2024csot,mohan2024progressive,kappeler2025bridging, yang2026flares}. However, these advances still rely heavily on large-scale 3D annotations, which remain expensive and difficult to obtain~\cite{yang2026collaborative}.

Self-supervised learning (SSL) therefore offers a scalable path to pretrain LiDAR representations without annotations~\cite{wu2025sonata,abdelsamad2026dos,min2023occupancy}, reducing reliance on expensive 3D labels and improving data efficiency in autonomous driving and robotics, where unlabeled sensor logs are abundant but dense annotation is costly. 
A recurring observation in LiDAR SSL is that pretrained representations transfer strongly to per-point tasks such as semantic and panoptic segmentation~\cite{abdelsamad2026dos, yang2026pointins}. Transfer to 3D object detection, however, is often substantially less effective. This gap is particularly relevant for autonomous driving, where reliable 3D detection is a key precursor for downstream planning and multi-object tracking pipelines~\cite{contreras2024survey}.

We analyze recent SSL methods and identify a shared limitation: the learning signal is predominantly defined on \textbf{measured LiDAR returns}, i.e., points captured from visible surfaces. Self-distillation approaches, such as Sonata~\cite{wu2025sonata} and DOS~\cite{abdelsamad2026dos}, explicitly enforce feature matching over scanned points only. Masked autoencoding variants, such as Occ-MAE~\cite{min2023occupancy} and NOMAE~\cite{abdelsamad2025nomae}, typically treat voxels with no point returns as empty, rather than modeling them as potentially occluded or unobserved space. Even recent methods that introduce instance-aware cues, e.g., PointINS~\cite{yang2026pointins}, still inherit this constraint: they are trained on visible returns and often define instance centroids using only the observed points, which can be biased under occlusion. As a result, pretraining emphasizes representations of observed surfaces, while the structure of \textbf{occluded, unobserved regions} is not explicitly incorporated into the objective, as shown in Figure~\ref{fig:empirical_gap}. This bias can be adequate for tasks defined on measured returns, but detection requires robustness to missing structure since object extent and localization must be inferred from incomplete measurements. 

In this work, we propose \textbf{\coolname{}}, an SSL framework that learns beyond visible surfaces by hallucinating latent features in \textbf{local neighborhoods} around discovered instances, including unobserved regions. \coolname{} applies an encoder to measured LiDAR returns, after inducing pseudo-occlusion by randomly masking returns, and retains standard encoder-level supervision with two complementary signals: \textbf{Softmaps} for semantics and \textbf{center-offset prediction} for instance discovery. To extend learning into unobserved space, we introduce an efficient instance voxel dilation procedure that expands each discovered instance into a neighborhood spanning both observed and no-return regions, and a lightweight neighborhood predictor that infers representations for sampled neighborhood voxels conditioned on encoded measured context. The predictor is supervised under the same Softmap and center-offset objectives, with targets for observed and masked voxels drawn from the teacher encoder and targets for unobserved voxels hallucinated by the teacher predictor. Since masked observed voxels and true no-return voxels both lack point evidence, learning to reverse pseudo-occlusion transfers naturally to genuinely unobserved regions, directly reducing occlusion bias.

We evaluate \coolname{} across several benchmarks and observe consistent improvements in downstream 3D detection, with the largest gains under sparse scans and limited labels. On nuScenes and Waymo, \coolname{} improves pretrained backbones out of the box when training only a lightweight detector, and further surpasses prior state of the art after full-parameter fine-tuning. On data-efficient benchmarks, it surpasses prior methods and matches full-label supervision with only 10\% labels. On nuScenes, we additionally verify that these improvements do not come at the expense of segmentation transferability, confirming the scalability of \coolname{} across diverse downstream tasks. To summarize, this work contains the following contributions:
\begin{itemize}[nosep]
    \item We characterize and analyze a systematic transfer gap in LiDAR SSL: pretrained representations that perform well on segmentation transfer substantially less effectively to 3D object detection, and we attribute this to a representational mismatch between sparse surface-aligned SSL objectives and the dense voxel space in which detection labels are annotated.
    \item We propose \textbf{\coolname{}}, an SSL framework that closes this gap by hallucinating latent features in local neighborhoods generated by instance voxel dilation, with an additional predictor and predictor-level supervision on generated voxels.
    \item We demonstrate consistent improvements in downstream 3D detection across standard LiDAR benchmarks, with especially strong gains under sparse sensing and limited label budgets, while preserving strong performance on other downstream tasks.
\end{itemize}
\section{Related Works}
\label{sec:related_works}

{\parskip=0pt
\noindent\textbf{LiDAR-based 3D Object Detection}: 
3D object detection from LiDAR point clouds has advanced rapidly with efficient voxel- and pillar-based representations. Earlier works such as VoxelNet~\cite{zhou2018voxelnet} and SECOND~\cite{yan2018second} discretize point clouds into regular 3D grids and apply sparse convolutions, while PointPillars~\cite{lang2019pointpillars} reduces computation by collapsing the vertical axis into pillar features. Anchor-free formulations such as CenterPoint~\cite{yin2021center} simplify detection by predicting object centers and offsets, yielding strong performance on standard benchmarks~\cite{caesar2020nuscenes, sun2020waymo, geiger2012kitti}. More recently, transformer-based detectors~\cite{wu2024ptv3, he2022voxelset, misra2021end, lang2024point, dong2022mssvt} have shown improved long-range interaction and global context modeling. In parallel with these architectural advances, end-to-end autonomous driving has gained momentum~\cite{hu2023uniad,zhang2025bridgead}. Yet explicit 3D detection remains a key component as it provides geometric structure and interpretable safety cues that are difficult to replace in planning-critical scenarios~\cite{contreras2024survey}. Despite these advances, these methods remain fully supervised and require large volumes of precisely annotated 3D bounding boxes, motivating self-supervised pretraining to leverage abundant unlabeled LiDAR data and reduce annotation dependency.}

{\parskip=2pt
\noindent\textbf{Point Cloud Self-supervised Learning (SSL)}: 
SSL has emerged as a prominent paradigm for learning robust 3D representations from LiDARs without human annotations.  Contrastive learning methods typically enforce representation consistency across augmented views and separate features of different samples, instances, or regions~\cite{xie2020pointcontrast,nunes2022segcontrast,nisar2025psa,yin2022proposalcontrast,zhang2021self}, while masked modeling and reconstruction-based methods pretrain encoders by recovering geometric information from masked inputs, such as occupancy or point coordinates~\cite{min2023occupancy,yang2023gdmae,tian2023geomae,hess2023masked}. More recent work extends this line by reconstructing occupancy only within local neighborhoods and at multiple scales, improving robustness on large-scale point clouds in the autonomous driving domain~\cite{abdelsamad2025nomae}. With the rapid adoption of Transformer-based backbones in various 3D perception tasks~\cite{wu2024ptv3,zhao2021point,he2022voxelset,lai2023spherical}, recent SSL approaches have also incorporated such architectures into point cloud representation learning~\cite{abdelsamad2026dos,wu2025sonata}. These methods typically employ a teacher–student framework with shared architectures, where the teacher predictions are regularized to maintain distributional sharpness and supervise the student on corrupted inputs. Despite effectiveness, all of these approaches share a critical limitation that they formulate pretraining objectives exclusively over the visible points, while ignoring occluded and sparse regions, where LiDAR fails to return. \coolname{} departs from this paradigm by explicitly targeting unobserved local neighborhoods as the primary objective during pretraining, encouraging the encoder to internalize object geometry beyond measured surfaces.}

{\parskip=2pt
\noindent\textbf{Point Cloud Completion}: 
The idea of inferring unobserved 3D geometry has been extensively studied in the context of point cloud completion, where the goal is to reconstruct a full object shape from a partial observation~\cite{yang2024tulip}. Methods such as PCN~\cite{yuan2018pcn} and FoldingNet~\cite{yang2018foldingnet} learn to map partial inputs to complete point clouds using encoder-decoder architectures trained on paired partial-complete data. Implicit representations and occupancy prediction networks~\cite{peng2020convolutional, chibane20ifnet,wang2025top} take a complementary approach, learning continuous signed distance or occupancy that can be queried at arbitrary 3D locations. Despite their ability to model unobserved geometry, these methods are fundamentally supervised where they rely on complete ground-truth shapes or bounding boxes, and are applied at inference time as standalone reconstruction tools. \coolname{} shares the intuition that modeling unobserved structure is valuable, but targets self-supervised representation learning on sparse LiDAR scans rather than supervised shape completion.}
\section{Method}
\label{sec:method}

Building on the completion intuition, we study why current LiDAR SSL still transfers weakly to 3D detection and show that this gap stems from a representation mismatch between surface-only pretraining and box-level supervision under occlusion. We then introduce \coolname{}, which extends self-distillation with dilated instance neighborhoods and a predictor-level distillation loss on non-visible voxels (masked and no-return) to encourage reasoning beyond observed returns.

\subsection{The SSL Transfer Gap to 3D Object Detection}

\label{subsec:analysis}

\subsubsection{Empirical Transfer Gap}
\label{subsubsec:transfer_gap}

We begin with a direct empirical observation. We evaluate two representative SSL methods on two downstream tasks, semantic segmentation and 3D object detection, on nuScenes~\cite{caesar2020nuscenes}. Using a probing protocol, we freeze the pretrained encoder and train only a lightweight task decoder. For segmentation, we use the standard lightweight decoder probe~\cite{wu2025sonata,yang2026pointins}. For detection, we attach a CenterPoint BEV backbone and head~\cite{yin2021center} directly after the 3D encoder. All methods use PTv3~\cite{wu2024ptv3} for architectural comparability. As shown in Table~\ref{tab:empirical_gap}, SSL methods transfer strongly to segmentation, achieving \emph{near-supervised} mIoU. For 3D detection, however, the same pretrained encoders show a substantial gap to the supervised baseline in both mAP and NDS. PointINS~\cite{yang2026pointins} introduces instance-aware cues to improve instance-localization learning and achieves a modest gain over DOS, however, the detection gap remains large. 
\begin{wraptable}{r}{0.47\textwidth}
  \centering
  \vspace{-1em}
  \caption{Empirical gap of existing SSL approaches between semantic segmentation and object detection.}
        \fontsize{8pt}{8pt}\selectfont
        \label{tab:empirical_gap}
        \begin{tabular}{l?c?cc}
            \toprule
            \multirow{2}{*}{\textbf{Method}} & 
            \multicolumn{1}{c?}{\textbf{Sem. Seg}} &  
            \multicolumn{2}{c}{\textbf{Obj. Det}} \\
            \cmidrule(lr){2-2} \cmidrule(lr){3-4} 
            & mIoU  & mAP & NDS \\
            \midrule
            PTv3 (sup.) &80.3 &63.8& 68.4\\
            DOS~\cite{abdelsamad2026dos} & 79.2  & 55.4 & 61.6 \\
            PointINS~\cite{yang2026pointins} & 80.0 & 56.7 & 62.5\\
            PointINS + Box Reg. & - & 56.2& 62.4\\
            \bottomrule
        \end{tabular}
  \vspace{-1.5em}
\end{wraptable}

Figure~\ref{fig:empirical_gap} illustrates the reason. Even when SSL correctly groups visible points into an instance, occlusion yields only a partial object view, so the pseudo-instance centroid is biased toward visible surfaces instead of the true geometric center. We also test whether adding box regression resolves this. As shown in the fourth row of Table~\ref{tab:empirical_gap}, extending PointINS with box fitting and a box regression head~\cite{nisar2025psa} does not improve over the baseline and slightly degrades performance. The reason is that the regression target is still derived from the same occluded point cluster, so geometric objectives defined only on visible points inherit the bias they aim to correct. This negative result leads to a key conclusion: instance-aware pretraining is necessary but not sufficient. The limitation is not missing geometric learning, but the incompleteness of the observed scene.
\subsubsection{Representation Mismatch in SSL}
We formalize this observation as a fundamental mismatch between how SSL pretraining and downstream detection define and use representations. Let $\mathcal{V}=\mathcal{V}^o\cup\mathcal{V}^u$ be the full scene voxel set, where $\mathcal{V}^o$ contains measured returns and $\mathcal{V}^u$ has no observations. In prior SSL, transformer encoders (e.g., PTv3~\cite{wu2024ptv3}) operate only on $\mathcal{V}^o$, producing no representation for $v\in\mathcal{V}^u$. This sparsity enables efficiency in large outdoor scenes. Accordingly, SSL objectives are naturally defined over occupied voxels $\mathcal{V}^o$, encouraging representations aligned with observed measurements. This works well for per-point tasks such as semantic segmentation, whose labels are also defined on $\mathcal{V}^o$ and do not require reasoning about unobserved regions.

Detection is annotated differently: boxes $\mathcal{B}_k$ specify the full object extent in $\mathcal{V}$, including partially occluded regions. During supervised training, the model must reason beyond sparse observed returns, effectively extrapolating sparse encoder features into unobserved space for object-level predictions. This creates a \textbf{representation mismatch} between SSL pretraining and detection fine-tuning. SSL constrains the encoder only on occupied voxels, whereas detection depends on representation quality over the full object extent. Thus, an encoder trained only with SSL may be valid on $\mathcal{V}^o$ yet inconsistent in unobserved regions $\mathcal{V}^u$. The issue is especially severe in outdoor LiDAR scenes with heavy occlusion. This exposes a core limitation of existing SSL for 3D detection and motivates pretraining that explicitly models object-level structure beyond observed measurements.

\subsection{\coolname{} Framework}
\label{sec:ghostpoint}
\begin{figure}[t]
\centering
\includegraphics[width=1\linewidth]{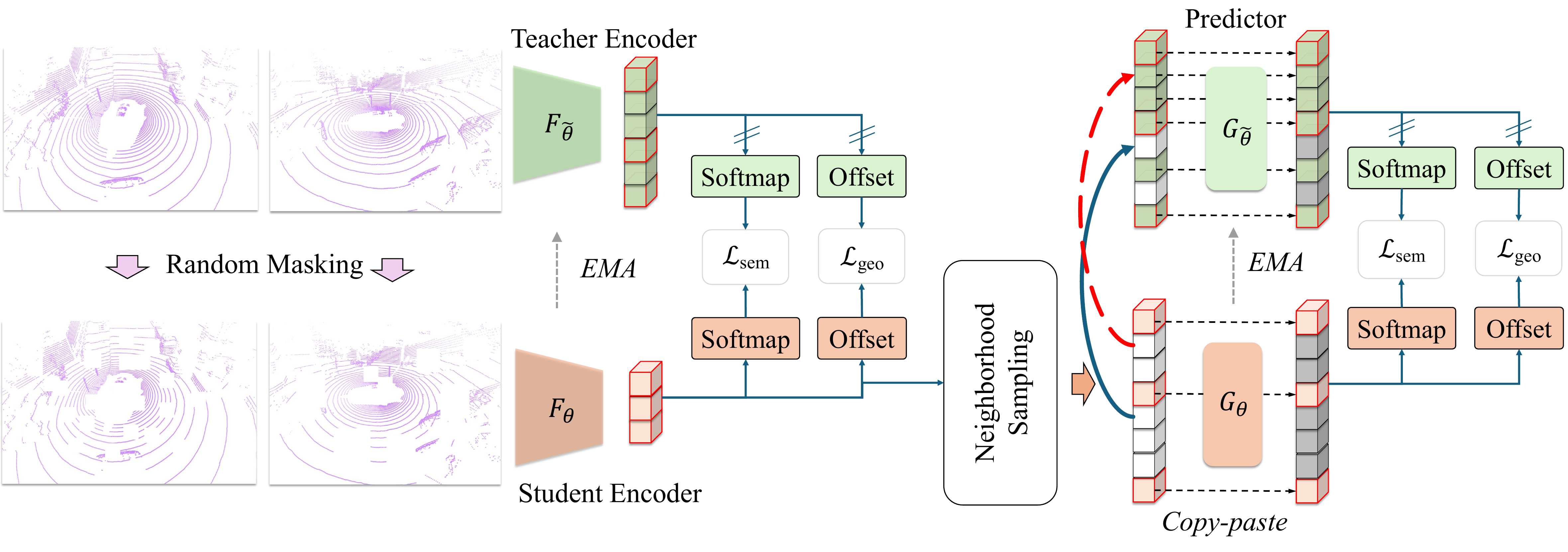}
\caption{\textbf{Overview of \coolname{}}: A scan is augmented into two views. The student encodes a masked version of each view $F_\theta$, while the teacher encoder $F_{\tilde{\theta}}$ (EMA) processes the unmasked views. Teacher features are used to discover instances. Discovered instances drive Neighborhood Sampling, which dilates their occupancy to form a neighborhood voxel set $\mathcal{Q}$ (visible, masked, and unobserved voxels). Teacher and student predictors $G_{\tilde{\theta}}$ and $G_\theta$ operate on tokens over $\mathcal{Q}$: the teacher hallucinates only at no-return voxels (overwriting occupied voxels with encoder features), while the student predicts over all non-visible voxels $\mathcal{Q}\setminus\mathcal{Q}_{\mathrm{vis}}$. Softmap and offset heads produce targets and predictions at two levels: visible voxels use encoder-level features, while the second-level supervision on non-visible voxels is described in Section~\ref{sec:ghostpoint}.}

\label{fig:method_overview}
\end{figure}

To address the mismatch identified above, we propose \coolname{}, which extends self-distillation~\cite{abdelsamad2026dos,yang2026pointins} with an explicit objective over unobserved regions as illustrated in Figure~\ref{fig:method_overview}.

Given an input point cloud, we generate two independently augmented views. For each augmented view $\mathcal{P}=\{(x_i,f_i)\}_{i=1}^N$, both student and teacher process the same view: the teacher receives the unmasked set $\mathcal{P}$, while the student receives its randomly masked subset $\mathcal{P}_{\mathrm{vis}}\subset\mathcal{P}$. Let $\mathcal{V}^o$ be the set of occupied voxels from voxelizing the teacher view $\mathcal{P}$, and let $\mathcal{V}^{\mathrm{vis}}\subseteq\mathcal{V}^o$ be the occupied voxels obtained by voxelizing the corresponding masked student input $\mathcal{P}_{\mathrm{vis}}$. The student encoder receives $\mathcal{V}^{\mathrm{vis}}$, while the teacher encoder receives $\mathcal{V}^o$, and the masked occupied voxels are $\mathcal{V}^{\mathrm{mask}}=\mathcal{V}^o\setminus\mathcal{V}^{\mathrm{vis}}$. Student and teacher share the same architecture, with teacher parameters updated as an EMA of the student. Both encoders output sparse per-voxel features. As in prior self-distillation~\cite{yang2026pointins}, Softmap and offset heads produce semantic assignments and center-offset predictions to form pseudo-instances and provide the base learning signal over matched visible voxels of the student. \coolname{} augments this setup with a prediction branch that extends supervision into occluded space, described next.

\begin{figure}[t]
\centering
\includegraphics[width=1\linewidth]{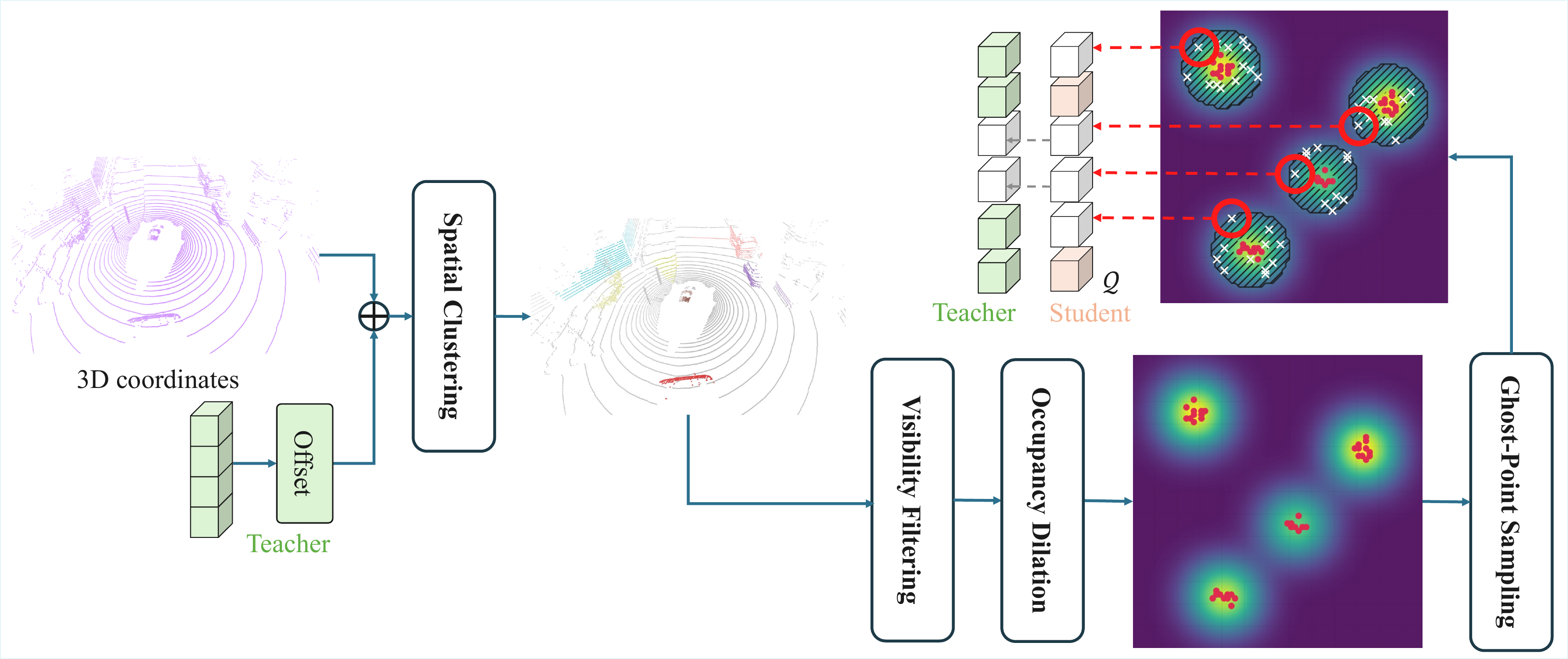}
\caption{\textbf{Neighborhood Sampling.} Teacher encoder features are passed to an offset head to predict center-offsets and cluster non-ground points into instance proposals. We drop proposals fully masked in the student view (visibility filtering), then dilate each remaining instance occupancy (kernel size $ks>1$) to form a neighborhood voxel set $\mathcal{Q}$ containing visible, masked, and newly activated no-return voxels. $\mathcal{Q}$ defines predictor queries and Softmap/offset supervision's scope for predictor-level distillation.}
\label{fig:neighborhood_sampling}
\end{figure}

\subsubsection{New Voxels Sampling}
The challenge of modeling unobserved regions without labels is knowing \emph{where} to place new queries in unoccupied space. Uniform sampling in free space is computationally prohibitive at outdoor scene scale and
yields mostly background locations with little structure to predict~\cite{abdelsamad2025nomae}. \coolname{}
instead generates queries \emph{near objects}: LiDAR returns are spatially clustered, and missing object
surfaces are most likely to occur adjacent to observed instance occupancy (e.g., due to occlusions or
no-return gaps), while voxels far from any instance are predominantly background. We implement this
instance-centric sampling in two steps: filtering student-visible instances and dilating their occupancy,
as illustrated in Figure~\ref{fig:neighborhood_sampling}.

In the first step, we drop pseudo-instances that are fully masked in the student view, keeping an instance if it contains at least one visible voxel. Let $\mathcal{V}^o_{\mathrm{inst}}\subseteq\mathcal{V}^o$ denote the union of occupied voxels from the remaining instances. We then dilate the binary occupancy grid of $\mathcal{V}^o_{\mathrm{inst}}$ with a 3D kernel of size $ks>1$ to obtain a candidate neighborhood set $\mathcal{Q}^{\mathrm{dil}}\supseteq\mathcal{V}^o_{\mathrm{inst}}$. For efficiency, we sample a fixed-size subset $\mathcal{Q}\subseteq\mathcal{Q}^{\mathrm{dil}}$ and use this sampled set for all subsequent predictor queries and losses. With respect to the student masking, $\mathcal{Q}$ decomposes into $\mathcal{Q}_{\mathrm{vis}}=\mathcal{Q}\cap\mathcal{V}^{\mathrm{vis}}$, $\mathcal{Q}_{\mathrm{mask}}=\mathcal{Q}\cap\mathcal{V}^{\mathrm{mask}}$, and $\mathcal{Q}_{\mathrm{unobs}}=\mathcal{Q}\setminus\mathcal{V}^o$.

\subsubsection{Token Initialization and Refinement}
The student predictor operates on the neighborhood voxel set $\mathcal{Q}$, which includes both occupied
voxels and newly activated no-return voxels. Tokens are available from the student encoder only on the
visible occupied subset $\mathcal{Q}_{\mathrm{vis}}=\mathcal{Q}\cap\mathcal{V}^{\mathrm{vis}}$. For all
remaining queried voxels $\mathcal{Q}\setminus\mathcal{Q}_{\mathrm{vis}}$ (including masked occupied voxels
$\mathcal{Q}_{\mathrm{mask}}$ and no-return voxels $\mathcal{Q}_{\mathrm{unobs}}$), the student must
initialize tokens explicitly before applying the predictor $G_\theta$. By default, we initialize each such
voxel $q$ by distance-weighted interpolation from its $k$ nearest occupied voxel neighbors in
$\mathcal{V}^{\mathrm{vis}}$ (we use $k=3$). Here, $\mathcal{N}_k(q)$ returns the $k$ nearest occupied
voxels $v\in\mathcal{V}^{\mathrm{vis}}$ to voxel $q$, and $c(\cdot)$ denotes a voxel's 3D center coordinate:
\begin{equation}
    \mathbf{f}_{q}=\sum_{v\in\mathcal{N}_k(q)} w_v\,\mathbf{f}_v,
    \qquad
    w_v=\frac{1/d_v}{\sum_{v'\in\mathcal{N}_k(q)} 1/d_{v'}},
    \label{eq:student_interp}
\end{equation}
where $d_v=\|c(v)-c(q)\|_2$ and $\mathbf{f}_v$ denotes the student encoder feature at visible occupied voxel
$v$. As an alternative, we also evaluate a simpler \emph{mask-token} initialization, where all voxels in
$\mathcal{Q}\setminus\mathcal{Q}_{\mathrm{vis}}$ share a learnable token $\mathbf{m}\in\mathbb{R}^d$.
The initialized tokens, together with encoder tokens on $\mathcal{Q}_{\mathrm{vis}}$, are then processed by
the lightweight predictor $G_\theta$ to aggregate information from visible voxels into the new tokens across $\mathcal{Q}$. To refine only newly queried tokens while using self-attention over all tokens for simplicity, we copy visible-token features from before the predictor to after the predictor (identity overwrite on $\mathcal{Q}_{\mathrm{vis}}$), and keep predictor outputs only on $\mathcal{Q}\setminus\mathcal{Q}_{\mathrm{vis}}$. Softmap and offset heads then map these features to semantic and geometric predictions over $\mathcal{Q}$.

\subsubsection{New Voxels Target Generation}

To supervise these predictions, we construct asymmetric \emph{teacher targets} on $\mathcal{Q}$ that are anchored on measured occupied voxels and hallucinate only in no-return space, following the same visible-token identity overwrite described above. For occupied voxels ($\mathcal{Q}\cap\mathcal{V}^o$, including $\mathcal{Q}_{\mathrm{mask}}$), we compute target Softmaps and offsets from the teacher \emph{encoder} representations, since they are directly supported by LiDAR returns. For unobserved voxels $\mathcal{Q}_{\mathrm{unobs}}$, we instead compute targets from the teacher \emph{predictor} output, providing context-based semantic and geometric predictions where no encoder token exists. Because the teacher predictor is an EMA of the student predictor and is conditioned on the full (unmasked) voxel set, these hallucinated targets become progressively stronger as training proceeds. We apply predictor-level distillation losses on $\mathcal{Q}\setminus\mathcal{Q}_{\mathrm{vis}}=\mathcal{Q}_{\mathrm{mask}}\cup\mathcal{Q}_{\mathrm{unobs}}$, thereby training masked occupied voxels and unobserved voxels under a unified objective and encouraging the same semantic and geometric completion behavior to transfer from pseudo-occluded to truly unobserved regions without any annotations.


\subsubsection{Training objectives}
\coolname{} applies distillation at two complementary stages: encoder-level supervision on voxels visible
to the student, and predictor-level supervision on non-visible voxels in the sampled neighborhood
(including masked and unobserved regions).

\paragraph{Encoder-level distillation.}
We apply standard self-distillation on visible occupied voxels $\mathcal{V}^{\mathrm{vis}}$.
Teacher and student encoder outputs are mapped to (i) Softmap distributions via the prototype head and
(ii) center-offset vectors via the offset head. The resulting semantic and geometric losses,
$\mathcal{L}_{\mathrm{sem,vis}}$ and $\mathcal{L}_{\mathrm{geo,vis}}$, are computed over
$\mathcal{V}^{\mathrm{vis}}$ following prior work~\cite{abdelsamad2026dos,yang2026pointins}.

\paragraph{Predictor-level distillation.}
Using the sampled neighborhood voxel set for the kept instances $\mathcal{Q}$ and its student-visible subset
$\mathcal{Q}_{\mathrm{vis}}=\mathcal{Q}\cap\mathcal{V}^{\mathrm{vis}}$, predictor-level supervision is applied on
$\mathcal{Q}_{\neg\mathrm{vis}}=\mathcal{Q}\setminus\mathcal{Q}_{\mathrm{vis}}$. We retain the same semantic (Softmap)
and geometric (offset) objectives used at the encoder level, following prior self-distillation findings
that these two signals are complementary for representation learning :
\begin{equation}
    \mathcal{L}_{\text{sem},\neg\mathrm{vis}} =
    \sum_{q \in \mathcal{Q}_{\neg\mathrm{vis}}}
    \mathrm{KL}\!\left(\pi^{G_{\tilde{\theta}}}(q)\;\|\;\pi^{G_\theta}(q)\right),
    \label{eq:sem}
\end{equation}
\begin{equation}
    \mathcal{L}_{\text{geo},\neg\mathrm{vis}} =
    \sum_{q \in \mathcal{Q}_{\neg\mathrm{vis}}}
    \left(
    \bigl\|\,\|\delta^{G_\theta}(q)\|-\|\tilde{\delta}^{G_{\tilde{\theta}}}(q)\|\,\bigr\|_1
    + 1 - \frac{\delta^{G_\theta}(q)\cdot\tilde{\delta}^{G_{\tilde{\theta}}}(q)}
    {\|\delta^{G_\theta}(q)\|\,\|\tilde{\delta}^{G_{\tilde{\theta}}}(q)\|}
    \right),
    \label{eq:geo}
\end{equation}
where $\pi^{G_{\tilde{\theta}}}(q)$ and $\pi^{G_\theta}(q)$ denote the Softmap distributions at voxel $q$,
and $\tilde{\delta}^{G_{\tilde{\theta}}}(q)$ and $\delta^{G_\theta}(q)$ denote the corresponding teacher and
student predictor offsets.
The full objective is
\begin{equation}
    \mathcal{L} =
    \mathcal{L}_{\text{sem,vis}} + \lambda\,\mathcal{L}_{\text{geo,vis}}
    + \mathcal{L}_{\text{sem},\neg\mathrm{vis}} + \lambda\,\mathcal{L}_{\text{geo},\neg\mathrm{vis}},
    \label{eq:total}
\end{equation}
with $\lambda$ balancing semantic and geometric terms.
\section{Experiments}
\label{sec:experiment}

Building on the \coolname{} design introduced in Section~\ref{sec:ghostpoint}, we evaluate how well the resulting pretrained representations transfer to 3D object detection. We first describe implementation details and benchmarks, then report main detection results under probing and full fine-tuning, followed by component ablations and additional analyses on label efficiency, segmentation transfer, cross-dataset transferability, and architectural transferability.

\subsection{Implementation Details}
We evaluate \coolname{} with two encoder backbones for $F_\theta$: PTv3~\cite{wu2024ptv3} and the sparse convolutional backbone of CenterPoint~\cite{yin2021center}. In both cases, we use the same lightweight predictor $G$, implemented as two PTv3-style transformer blocks. Training uses a two-stage warmup before joint optimization: for the first 10\% of epochs we train only the encoder Softmap branch to stabilize representations; for the next 10\% we activate the encoder offset branch and predictor branches, detaching their gradients from the student encoder for stability. We then optimize all components end-to-end. Performance is robust to these hyperparameters as long as the warmup order is preserved (see Appendix). Unless specified otherwise, we set $ks=5$ for occupancy dilation and $\lambda=0.1$ for loss weighting. For downstream evaluation, the pretrained encoder serves as the backbone of a CenterPoint detector~\cite{yin2021center}. Additional details are provided in the Appendix.
\subsection{Dataset}
We evaluate \coolname{} on two widely adopted LiDAR 3D object detection benchmarks. \textbf{nuScenes}~\cite{caesar2020nuscenes} is a large-scale autonomous driving dataset comprising 700 training, 150 validation, and 150 test scenes captured with a 32-beam LiDAR sensor. It provides 1.4M annotated 3D bounding boxes across 10 object categories, with performance measured by mean Average Precision (mAP) and the nuScenes Detection Score 
(NDS). \textbf{Waymo Open Dataset}~\cite{sun2020waymo} is a large-scale benchmark containing 798 training and 202 validation sequences, annotated across 3 object categories. Following prior works~\cite{nisar2025psa, zhu2025self}, we use 20\% of full training set for pretraining and report Average Precision (AP) at Level 2 difficulty.

\begin{table}[t]

\centering
\caption{\textbf{Main results} of SSL pretraining for LiDAR 3D object detection with PTv3 as backbone. \emph{probe}: freeze the backbone $F_\theta$ and train the remaining detector components from scratch; \emph{ft}: end-to-end fine-tuning. Best results in each setting are in \textbf{bold}.}
\vspace{-1em}

\fontsize{8pt}{8pt}\selectfont
\setlength{\tabcolsep}{7pt}
\begin{tabular}{l|cc|cccc}
\toprule
\multirow{2}{*}{\textbf{Method}}& \multicolumn{2}{c|}{\textbf{nuScenes}} & \multicolumn{4}{c}{\textbf{Waymo (L2 mAP)}} \\
\cmidrule(lr){2-3} \cmidrule(lr){4-7}
 & mAP & NDS &  Vehicle& Pedestrian & Cyclist & mAP\\
\midrule
 \rowcolor{gray!12} PTv3 (sup.)& 63.8 & 68.4& 69.5 & 67.5 & 66.7 & 67.9 \\
 \midrule
 PSA~\cite{nisar2025psa} (prob.) & 41.3 & 52.8 & 45.6 & 42.4& 40.8 & 42.4\\
 SONATA~\cite{wu2025sonata} (prob.)& 44.6 & 55.0 & 51.2 & 46.2& 44.9 &47.4\\
 NOMAE~\cite{abdelsamad2025nomae} (prob.)&  53.5 & 60.1 & 55.3 & 51.5 & 50.3 & 52.4 \\
 DOS~\cite{abdelsamad2026dos} (prob.)& 55.4 & 61.6 & 59.4 &56.6 &55.2 & 57.1\\
 PointINS~\cite{yang2026pointins} (prob.) & 56.7 & 62.5 & 60.2& 56.6 &  55.7 & 57.5\\
\rowcolor{blue!7}\coolname{} (prob.) & \textbf{59.5} & \textbf{64.2} & \textbf{62.3} & \textbf{59.2}& \textbf{58.5}& \textbf{60.0}\\
\midrule
PSA~\cite{nisar2025psa} (ft.) &63.9  & 68.9 & 69.9& 67.5 &67.0 &68.1 \\
 SONATA~\cite{wu2025sonata} (ft.)& 64.2 & 69.0 & 70.8 &67.9 &67.0 & 68.6 \\
 NOMAE~\cite{abdelsamad2025nomae} (ft.)& 65.8 & 69.8 &71.4 & 68.5& 68.5& 69.5\\
 DOS~\cite{abdelsamad2026dos} (ft.)& 65.5& 69.7 & 71.4& 68.2 & 68.0& 69.2 \\
 PointINS~\cite{yang2026pointins} (ft.) & 66.3 & 70.1 & 71.6 & 68.5& 68.3& 69.5\\
 \rowcolor{blue!7}\coolname{} (ft.) & \textbf{67.5} & \textbf{71.2} & \textbf{72.0} & \textbf{69.2} & \textbf{69.0} &\textbf{70.1}  \\
\bottomrule
\end{tabular}
\label{tab:main_results}
\end{table}

\subsection{Main Results}
\begin{figure}[t]
\centering
\includegraphics[width=1\linewidth]{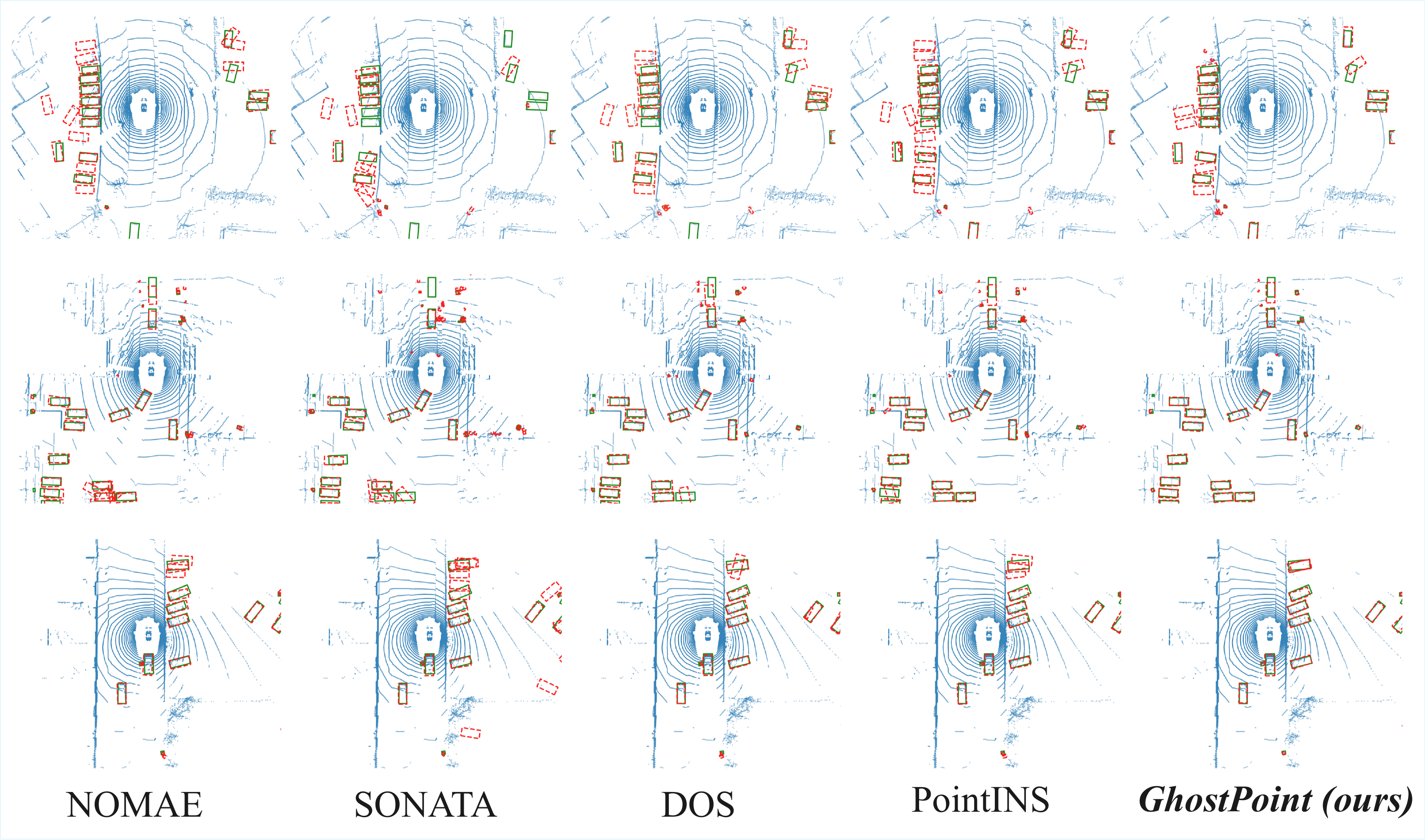}
\caption{Qualitative results on the nuScenes dataset~\cite{caesar2020nuscenes}. All self-supervised models are evaluated under the decoder-probing protocol. \textcolor{green}{Green} boxes denote ground-truth annotations and \textcolor{red}{red} boxes denote predictions. \coolname{} produces predictions that align more closely with ground-truth boxes in both center localization and spatial extent, particularly for partially observed and occluded objects at mid-to-long range.}
\label{fig:qualitative}
\end{figure}

Table~\ref{tab:main_results} compares \coolname{} against recent SSL methods under two evaluation protocols: decoder probing, where the pretrained PTv3 encoder is frozen and other CenterPoint components are trained from scratch, and full fine-tuning, where the encoder is initialized from pretrained weights and the full detector is optimized end-to-end. \coolname{} consistently achieves the best performance. Under decoder probing, \coolname{} surpasses the second-best SSL method by 2.8 mAP and 1.7 NDS on nuScenes, and by 2.5 mAP on Waymo. Under full fine-tuning, gains are further amplified, with \coolname{} reaching 67.5 mAP and 71.2 NDS on nuScenes and 70.1 mAP on Waymo, surpassing the supervised baseline by 3.7 mAP on nuScenes. These results demonstrate that explicitly modeling unobserved object geometry during pretraining yields representations that are substantially better aligned with 3D object detection. Additionally, Figure~\ref{fig:qualitative} highlights these gains: in the first row, \coolname{} is the only method that correctly localizes the partially observed car at the top right. In the second row, it reduces false positives and improves center/orientation accuracy, although it can still hallucinate occasional boxes in empty space. In the third row, it is the only method that consistently produces a single prediction per object with the correct orientation.

\subsection{Influence of Key Components} 
\begin{figure}[t]
\centering
\begin{minipage}{0.52\textwidth}
    \centering
    \captionof{table}{\textbf{Ablation study on \coolname{}'s key components}. We evaluate on nuScenes~\cite{caesar2020nuscenes} under decoder probing protocol. $^\star$: no predictor, features of  newly sampled voxels are interpolated from observed nearest neighbors.
    }
    \fontsize{8pt}{8pt}\selectfont
    \begin{tabular}{l|cc}
    \toprule
     \textbf{Model} & mAP & NDS \\ 
    \midrule
    Baseline~\cite{yang2026pointins} & 56.7& 62.5\\
    + $^\star$Neighborhood Sampling & 57.3 & 62.7  \\
    + Predictor (no warmup) & 57.2 & 63.0 \\
    + Predictor (one-stage warmup) & 58.5 & 63.6 \\
    + Predictor (two-stage warmup) & 59.5 & 64.2 \\
    \textcolor{lightgray}{+Box Regression}& 59.1 & 63.9\\
    \bottomrule
    \end{tabular}
    \label{tab:ablation_study}
    \end{minipage}
    \hfill
    \begin{minipage}{0.43\textwidth}
        \centering
        \includegraphics[width=\linewidth]{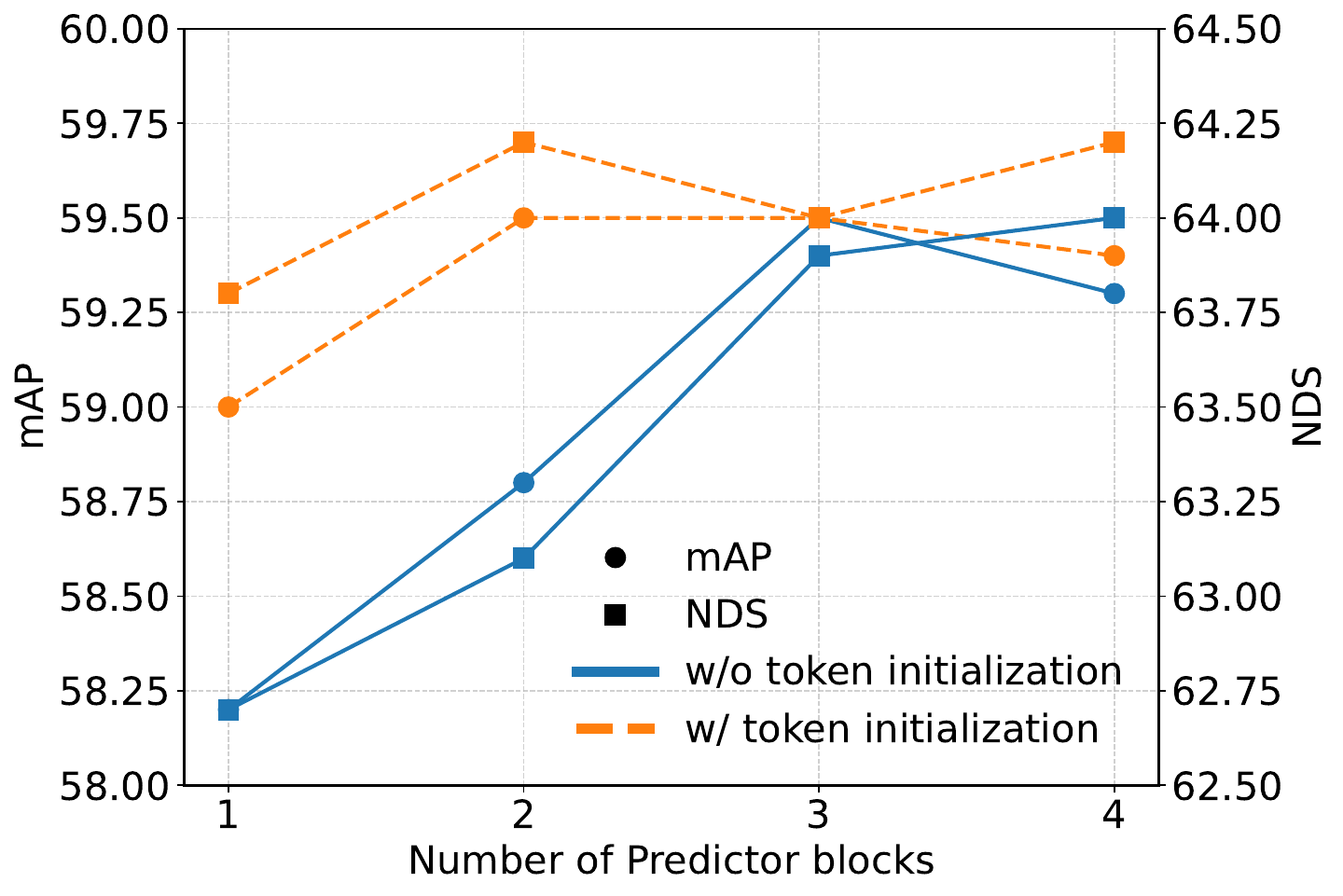}
        \caption{\textbf{Effect of predictor depth and token initialization}. We ablate number of blocks in $G_\theta$ and KNN feature interpolation for teacher token initialization.}
        \label{fig:block_init_ablation}
    \end{minipage}
\vspace{-5mm}
\end{figure}

Table~\ref{tab:ablation_study} ablates the key components of \coolname{}. Starting from the baseline~\cite{yang2026pointins}, we first add the Neighborhood Sampling module with direct distance-weighted KNN interpolation, assigning features to ghost positions from their nearest observed neighbors without any additional predictor. This already yields a consistent improvement, which confirms that extending the supervision signal into occluded regions is beneficial for SSL even without additional model capacity. Adding the predictor with the warmup schedule produces a substantially larger 
gain. This improvement is not merely a capacity effect: the predictor enables active context propagation between observed and unobserved positions, requiring the encoder to produce features that are aware of object geometry beyond directly measured surfaces. Finally, consistent with the analysis in Section~\ref{subsec:analysis}, augmenting \coolname{} with an additional box 
regression target leads to a performance drop. This observation reinforces our conclusion that additional geometric supervision cannot resolve the representational mismatch in SSL. 

Figure~\ref{fig:block_init_ablation} reports the effect of predictor depth and token initialization scheme. Without KNN-based initialization of newly sampled tokens, performance increases consistently with predictor depth up to 3 blocks, then plateaus. This suggests that additional capacity helps propagate context between observed and ghost positions but yields diminishing returns beyond a moderate depth. With the token initialization, performance is consistently higher and remains stable from 2 blocks onward, which indicates that providing the teacher with spatially coherent ghost initializations reduces the burden on the predictor and makes training less sensitive to predictor depth. Based on these results, we adopt 2 predictor blocks as the default, balancing performance and computational cost. Additional ablations on other hyperparameters are presented in the Appendix.

\begin{wraptable}{r}{0.47\textwidth}
  \centering
  \vspace{-2em}
  \caption{Disentangling hallucination from regularization 
  on nuScenes (decoder probing).}
  \label{tab:disentangle}
  \small
  {\fontsize{8}{10}\selectfont
  \begin{tabular}{lcc}
    \toprule
    Model & mAP & NDS \\
    \midrule
    PointINS                              & 56.7 & 62.5 \\
    +~No hallucination                    & 57.3 & 63.1 \\
    +~Zero targets                        & 57.6 & 63.3 \\
    +~Swap masked $\to$ hallucinated      & 59.4 & 64.1 \\
    \textit{GhostPoint} (ours)            & \textbf{59.5} & \textbf{64.2} \\
    \bottomrule
  \end{tabular}
  }
  \vspace{-2em}
\end{wraptable}

\PAR{Disentangling hallucination from regularization.} Table~\ref{tab:disentangle} isolates the source of gains through three controlled ablations. First, restricting the predictor to masked observed points (\emph{No hallucination}) slightly improves over PointINS (+0.6 mAP / +0.6 NDS), suggesting mild regularization from predictor capacity. Second, keeping ghost-point neighborhoods but replacing their targets with zero offsets and uniform Softmaps (\emph{Zero targets}) yields only marginal further change. This shows that ghost-point locations alone contribute little. Third, swapping the same number of masked points with hallucinated points (\emph{Swap}) matches full \coolname{}, ruling out any supervision-count advantage. Together, these results attribute the gains specifically to \textbf{informative hallucination targets} in object-adjacent unobserved regions, rather than to predictor regularization or extra supervised positions.

\subsection{Visualizations}
\begin{figure}[t]
\centering
\begin{subfigure}{0.24\linewidth}
    \includegraphics[width=\columnwidth]{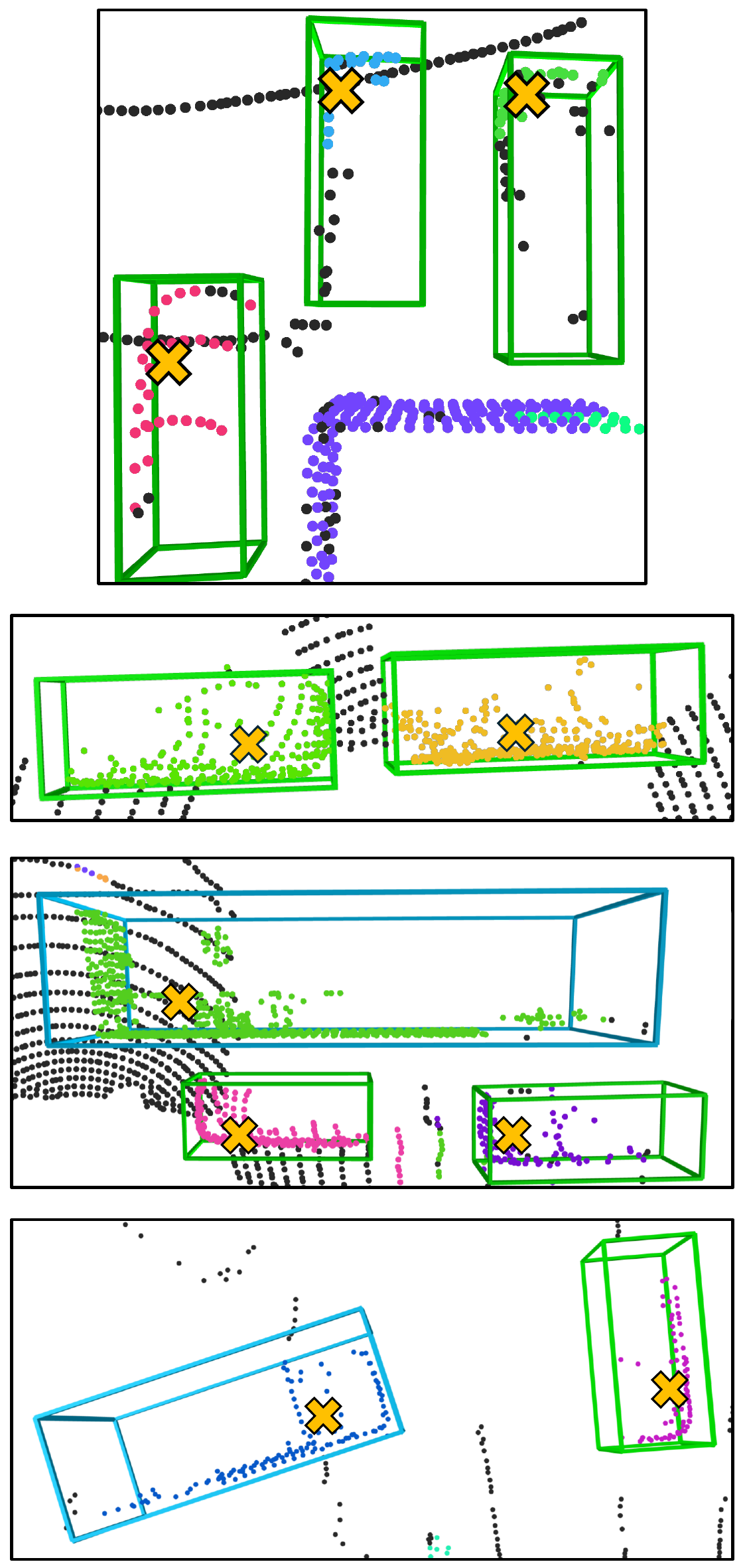}
    \caption{orig. Instance}
    \label{subfig:inst_nohalluc}
\end{subfigure}
\begin{subfigure}{0.24\linewidth}
    \includegraphics[width=\columnwidth]{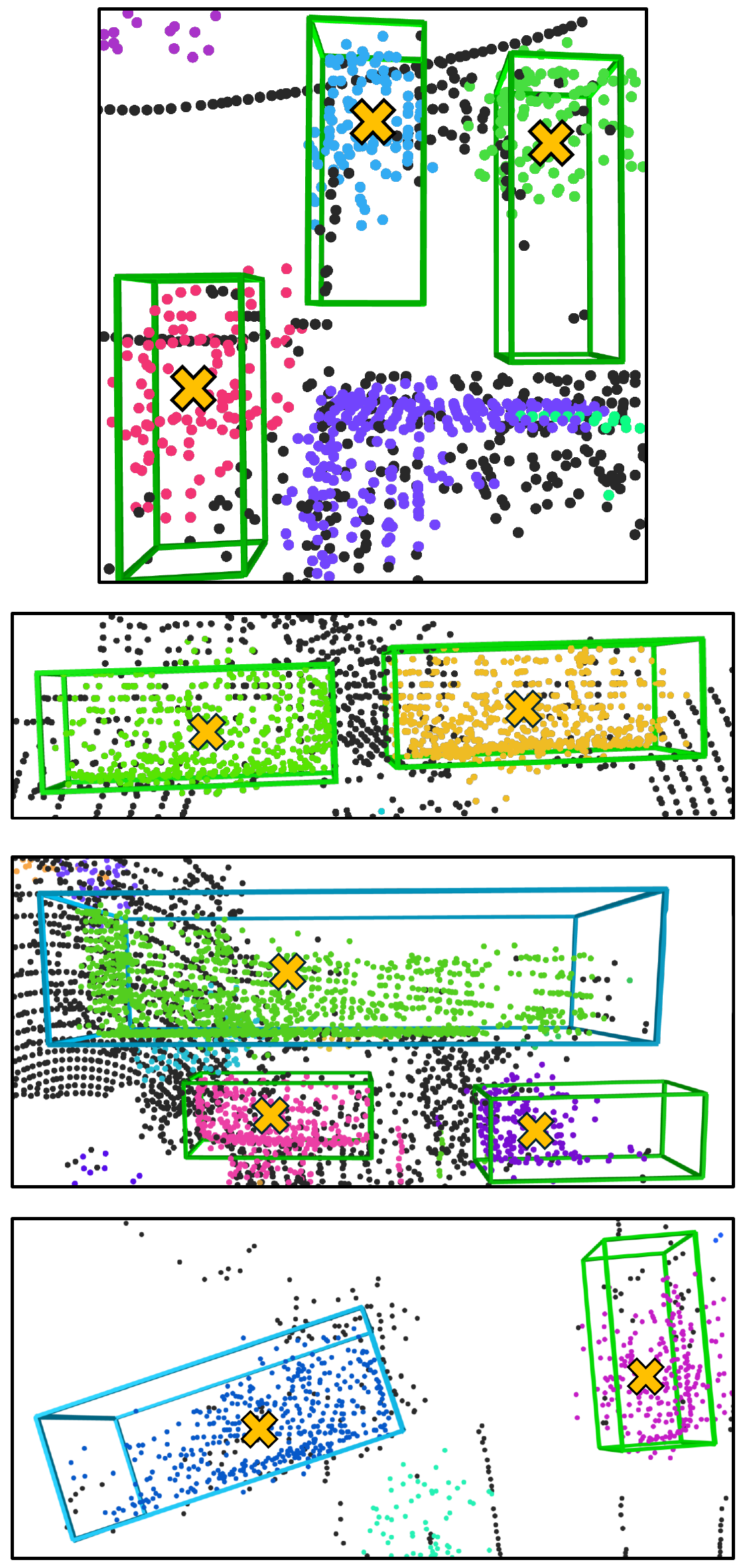}
    \caption{halluc. Instance}
    \label{subfig:inst_halluc}
\end{subfigure}
\begin{subfigure}{0.24\linewidth}
    \includegraphics[width=\columnwidth]{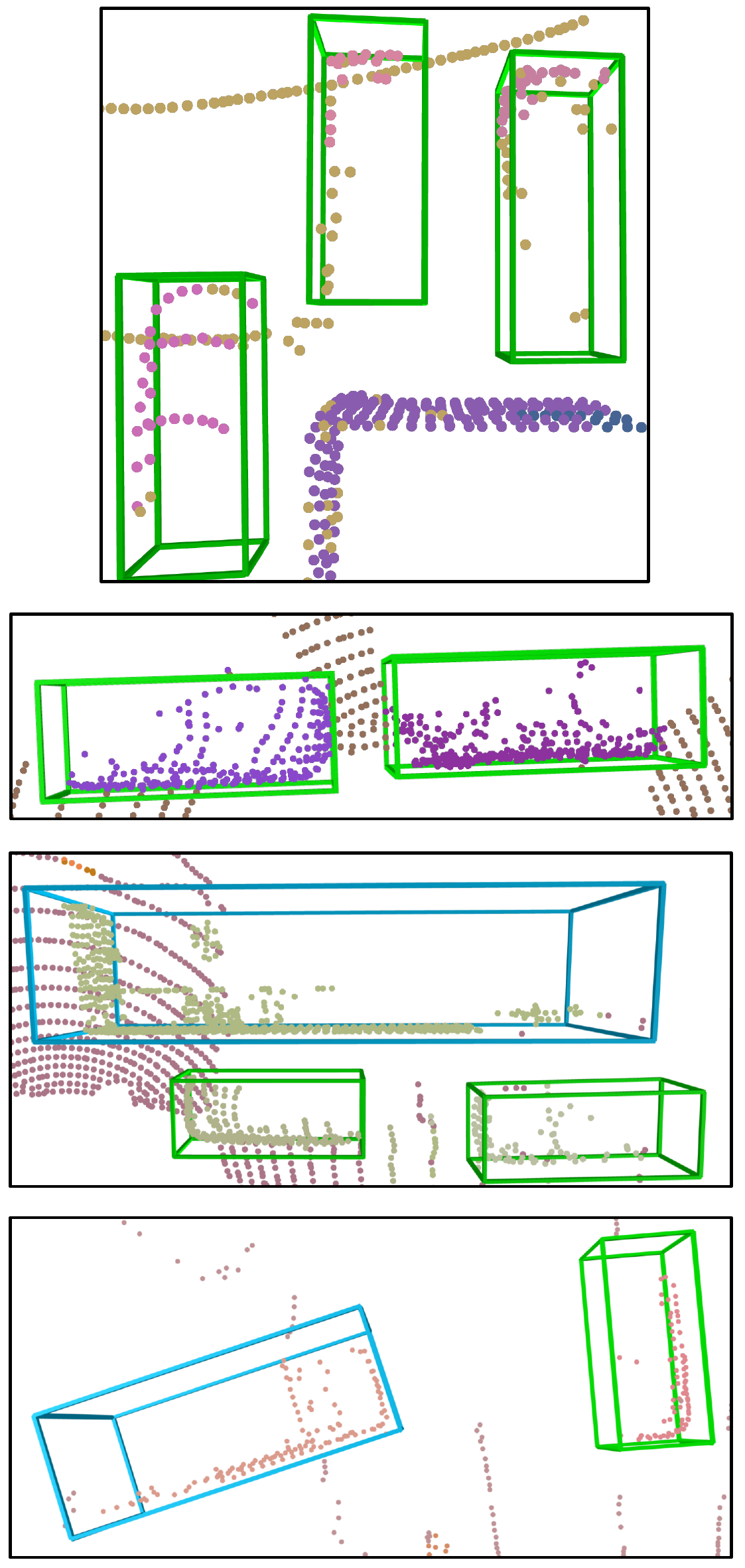}
    \caption{orig. PCA}
    \label{subfig:pca_nohalluc}
\end{subfigure}
\begin{subfigure}{0.24\linewidth}
    \includegraphics[width=\columnwidth]{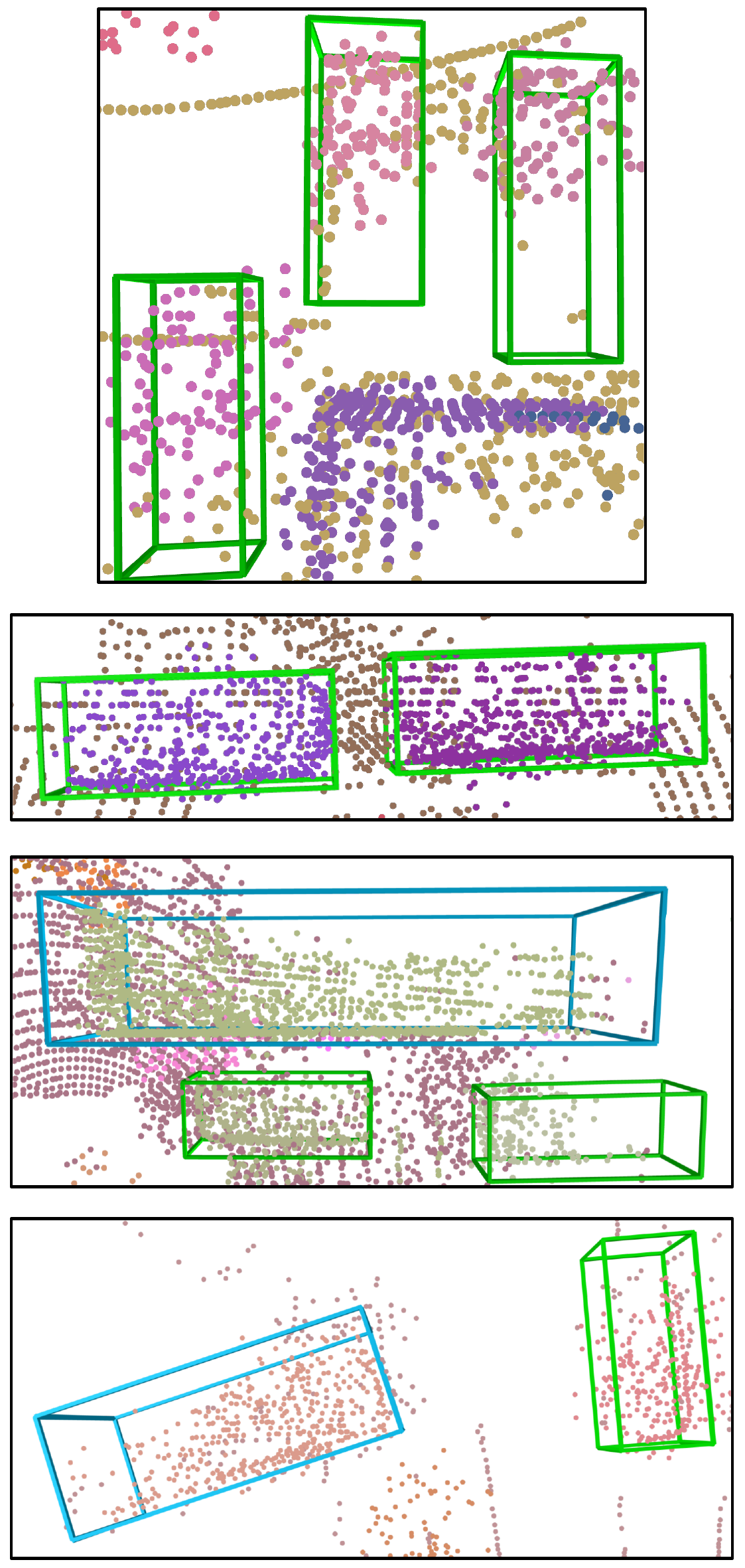}
    \caption{halluc. PCA}
    \label{subfig:pca_halluc}
\end{subfigure}
\caption{Visualizations of original and hallucinated point clouds. Ground-truth bounding boxes are shown for reference. (a)/(b): pseudo instances obtained by clustering predicted offsets; (c)/(d): PCA projections of point features.}
\vspace{-2mm}
\label{fig:inst_pca}
\end{figure}

To further analyze what is learned within \coolname{}, we provide visualizations in Fig.~\ref{fig:inst_pca}. We apply clustering to predicted center offsets and visualize the resulting pseudo-instance assignments in color, examining whether hallucinated points meaningfully extend object geometry into occluded regions. We additionally visualize PCA-projected features to assess the semantic coherence of hallucinated points. Both views consistently show that \coolname{} produces object-consistent representations rather than arbitrary fillers, and that the hallucinated neighborhoods reduce visible-surface bias under occlusion.

Nonetheless, the visualizations also expose clear failure modes. When objects are extremely sparse due to severe occlusion, as in the first example, the hallucination fails to recover the complete object shape. Additionally, hallucination can be sensitive to background outliers, leading to blurred or incorrectly extended instance boundaries. 

\subsection{Additional Results}

\subsubsection{Label Efficiency}
\begin{figure}[t]

\centering
\vspace{-1em}
\begin{minipage}{0.6\textwidth}
  \captionof{table}{ \textbf{Label efficiency results on nuScenes.}
  }
  \setlength{\tabcolsep}{3pt}
  \fontsize{8pt}{8pt}\selectfont
  \begin{tabular}{l|cc|cc|cc}
  \toprule
  \multirow{2}{*}{\textbf{Method}} & \multicolumn{2}{c|}{\textbf{0.1\%}} & \multicolumn{2}{c|}{\textbf{1\%}}& \multicolumn{2}{c}{\textbf{10\%}}\\
  \cmidrule(lr){2-3} \cmidrule(lr){4-5}  \cmidrule(lr){6-7} 
    &  mAP & NDS & mAP & NDS & mAP & NDS\\
  \midrule
  \rowcolor{gray!12} PTv3 (sup.) & 18.1 &35.3 & 37.3& 48.0 &57.9 & 65.0 \\
  \midrule
  PSA~\cite{nisar2025psa}    & 25.2&42.8 & 41.0& 51.9& 58.0& 65.2 \\
  SONATA~\cite{wu2025sonata}  & 26.4& 44.3& 43.2& 53.6& 58.4& 65.8\\
  NOMAE~\cite{abdelsamad2025nomae}  & 30.5& 46.8&45.7 &56.0 &59.7 &66.6 \\
  DOS~\cite{abdelsamad2026dos}  &34.0 &49.1 & 49.0& 58.0 & 61.4& 67.4 \\
  PointINS~\cite{yang2026pointins}  & 34.8& 50.8& 50.8& 60.2&62.1 &67.8 \\
  \rowcolor{blue!7}\coolname{} & \textbf{36.9} & \textbf{52.1}&  \textbf{53.1} & \textbf{62.5} &\textbf{63.5}& \textbf{68.8} \\
  \bottomrule
  \end{tabular}
  \label{tab:nuscenes_efficient}
\end{minipage}
\hfill
\begin{minipage}{0.35\textwidth}
    \setlength{\tabcolsep}{5pt}
    \captionof{table}{\textbf{\textbf{Wy $\rightarrow$ nS} Transfer Probing}
    }
    \begin{tabular}{l|cc}
    \toprule
    \textbf{Method}& mAP & NDS \\
    \midrule
    PSA~\cite{nisar2025psa}  & 28.2 &43.9  \\
    SONATA~\cite{wu2025sonata} & 32.2 & 47.3\\
    NOMAE~\cite{abdelsamad2025nomae} & 35.6 & 50.5\\
    DOS~\cite{abdelsamad2026dos} & 41.0 & 52.3 \\
    PointINS~\cite{yang2026pointins}  & 41.6 & 52.4 \\
    \rowcolor{blue!7}\coolname{} & \textbf{43.4} & \textbf{54.1} \\
    \bottomrule
    \end{tabular}
    \label{tab:transfer_probing}
    \end{minipage}
    \vspace{-1em}
\end{figure}

We evaluate \coolname{} on label-efficient benchmarks on both datasets. Results are shown in Table~\ref{tab:nuscenes_efficient} and Table~\ref{tab:waymo_efficient}. \coolname{} consistently outperforms all prior SSL methods across every annotation regime on both datasets. On nuScenes, with only 0.1\% of labels, \coolname{} achieves 36.9 mAP and 52.1 NDS, surpassing the second-best method by 2.1 mAP and 1.3 NDS. At 1\% labels, the margin widens to 2.3 mAP and 2.3 NDS over PointINS, and at 10\% labels \coolname{} reaches 63.5 mAP and 68.8 NDS. On Waymo, \coolname{} similarly leads across all categories at both 0.1\% and 1\% labels, with particularly strong gains on pedestrian and cyclist detection where occlusion is most severe. 

\begin{table}[t]

\centering
\vspace{0.8em}
\caption{\textbf{Label efficiency results on Waymo.}
}
\vspace{-1.0em}
\fontsize{8pt}{8pt}\selectfont
\setlength{\tabcolsep}{2.5pt}
\begin{tabular}{l|cccc|cccc}
\toprule
\multirow{2}{*}{\textbf{Method}}& \multicolumn{4}{c|}{\textbf{\textbf{0.1\%}}} & \multicolumn{4}{c}{\textbf{\textbf{1\%}}} \\
\cmidrule(lr){2-5} \cmidrule(lr){6-9}
 & Vehicle& Pedestrian & Cyclist & mAP &  Vehicle& Pedestrian & Cyclist & mAP\\
\midrule
 \rowcolor{gray!12} PTv3 (sup.)& 32.1 & 15.2 & 12.6 & 20.0 & 42.1 & 27.2 & 33.5 & 34.3  \\
  NOMAE~\cite{abdelsamad2025nomae} & 34.8  & 20.9 & 18.8 & 24.8& 43.6& 29.8 &35.0 & 36.1 \\
  DOS~\cite{abdelsamad2026dos} & 38.3 & 23.2 & 20.0 & 27.2 & 45.1& 33.2&  36.5& 38.3\\
PointINS~\cite{yang2026pointins} &  39.0 & 22.8& 19.6 & 27.1 &46.5 & 34.3&  37.2& 39.3\\
\rowcolor{blue!7}\coolname{} & \textbf{40.5} & \textbf{23.5}&\textbf{21.6} & \textbf{28.5}& \textbf{47.2}& \textbf{36.8}& \textbf{39.0}&\textbf{41.0} \\

\bottomrule
\end{tabular}
\label{tab:waymo_efficient}
\end{table}

\subsubsection{Segmentation Performance}
To verify that \coolname{} does not sacrifice per-point transferability in favor of detection, we evaluate semantic and panoptic segmentation on nuScenes under a linear-probing protocol (Table~\ref{tab:segmentation}). On semantic segmentation, \coolname{} reaches \textbf{74.2} mIoU, essentially matching the strongest SSL baseline (PointINS: \textbf{74.4}), which indicates that surface-level feature quality is preserved. On panoptic segmentation, \coolname{} achieves the best SSL performance with \textbf{62.8} PQ, improving over PointINS by \textbf{+0.6} and over DOS by \textbf{+5.4}; it also attains the top SQ/RQ among SSL methods (\textbf{84.6}/\textbf{73.2}). Overall, these results show that \coolname{} strengthens object-level transfer for detection while maintaining strong per-point semantics.
\begin{figure}[t]
    \centering
    \vspace{-0.7em}
    \begin{minipage}{0.5\textwidth}
    \captionof{table}{\textbf{Architectural transferability on Waymo (L2 mAP).} Replacing PTv3 with the CenterPoint sparse CNN encoder, distillation from a frozen \coolname{} PTv3 teacher consistently outperforms training the CNN directly (probing and fine-tuning).}
    \fontsize{8pt}{8pt}\selectfont
    \setlength{\tabcolsep}{4pt}
    \begin{tabular}{l|cccc}
    \toprule
    \textbf{SPUNet}& Veh. & Ped. & Cyc. & mAP \\ 
    \midrule
    From scratch & 64.9 & 62.3 & 66.3 & 64.5\\
    Pretrain (prob.) & 48.5& 43.4& 50.2& 48.0 \\
    Distill (prob.) & 58.3 & 54.7& 59.0 & 57.3\\ 
    Pretrain (ft.) & 65.8 & 63.5 &  66.9 & 65.4\\ 
    Distill (ft.) & \textbf{68.8} & \textbf{64.8} &  \textbf{68.0} & \textbf{67.2}\\ 
    \bottomrule
    \end{tabular}
    \label{tab:spunet}
    \end{minipage}
    \hfill
    \begin{minipage}{0.46\textwidth}
     \captionof{table}{\textbf{Linear probing performance on nuScenes Semantic and Panoptic Segmentation.}
    }
    \fontsize{8pt}{8pt}\selectfont
    \setlength{\tabcolsep}{2.5pt}
    \begin{tabular}{l|c|ccc}
    \toprule
    \multirow{2}{*}{\textbf{Method}}& \multicolumn{1}{c|}{\textbf{Sem.}} & \multicolumn{3}{c}{\textbf{Pan.}}\\
    \cmidrule(lr){2-2} \cmidrule(lr){3-5} 
    &  mIoU & PQ & SQ & RQ\\
    \midrule
     \rowcolor{gray!12} PTv3 (sup.) & 80.3 & 69.9& 86.3& 80.5\\
    \midrule
    PSA~\cite{nisar2025psa}  & 44.5& 30.1&73.9 &38.6 \\
    SONATA~\cite{wu2025sonata} & 59.2& 50.7& 79.8& 61.6\\
    NOMAE~\cite{abdelsamad2025nomae}& 64.7& 45.5 &77.0 &56.4 \\
    DOS~\cite{abdelsamad2026dos} & 74.1& 57.4 & 82.8& 68.5 \\
    PointINS~\cite{yang2026pointins} & \textbf{74.4} & 62.2& 84.5 &72.8  \\
    \rowcolor{blue!7}\coolname{} & 74.2 & \textbf{62.8} & \textbf{84.6} &\textbf{73.2}  \\
    \bottomrule
    \end{tabular}
    
    \label{tab:segmentation}
    \end{minipage}
    \vspace{-1em}
\end{figure}
\subsubsection{Cross-dataset Transferability}

To assess cross-domain generalization, we pretrain on Waymo and transfer to nuScenes under the decoder-probing protocol. As shown in Tab.~\ref{tab:transfer_probing}, \coolname{} achieves the best cross-dataset performance with \textbf{43.4} mAP and \textbf{54.1} NDS. This improves over the strongest prior SSL baseline (PointINS: 41.6 mAP, 52.4 NDS) by \textbf{+1.8} mAP and \textbf{+1.7} NDS, and over DOS (41.0 mAP, 52.3 NDS) by \textbf{+2.4} mAP and \textbf{+1.8} NDS, indicating better robustness to domain-specific sensor characteristics and scene layouts.

\subsubsection{Architectural Transferability}
Table~\ref{tab:spunet} reports \coolname{} results on \textbf{Waymo} when replacing the PTv3 encoder with the \textbf{convolutional encoder from CenterPoint}. Training \coolname{} directly with this CNN backbone transfers less effectively than with a transformer: under probing, the CNN variant reaches only \textbf{48.0} L2 mAP overall, compared to \textbf{57.3} when distilling from a frozen \coolname{} PTv3 teacher (\textbf{PTv3}$\rightarrow$\textbf{CNN}). With full fine-tuning, distillation further improves the CNN student to \textbf{67.2} overall, outperforming both the CNN trained with \coolname{} directly (\textbf{65.4}) and the supervised CNN baseline (\textbf{64.5}). This suggests \coolname{} pretrained PTv3 features are broadly useful beyond the PTv3 architecture, offering a practical path to transferring object-aware representations into lightweight CNN backbones where transformer encoders may be computationally prohibitive. This trend also aligns with observations in earlier work that pretraining a transformer and distilling to a CNN can outperform pretraining the CNN directly~\cite{abdelsamad2026dos}.
\section{Conclusion}
\label{sec:conclusion}
We presented \coolname{}, which addresses a key limitation of LiDAR SSL: objectives defined primarily on measured returns leave occluded and no-return regions largely unconstrained, contributing to a systematic transfer gap to 3D object detection. We attributed this gap to a representation mismatch between surface-aligned SSL supervision and detection labels defined over full object extents. \coolname{} mitigates this mismatch by generating instance-centric neighborhood voxels via instance voxel dilation and training a predictor to infer ghost representations from observed context. A predictor-level supervision scheme matches student predictions to teacher targets on non-visible neighborhood voxels, promoting features that support reasoning beyond visible surfaces. Experiments on nuScenes and Waymo show consistent improvements across evaluation protocols and label budgets, while maintaining strong transfer to segmentation. Our work has two limitations: under probing, performance still trails fully supervised training, and we have not evaluated \coolname{} on other sparse sensing modalities affected by occlusion (e.g., radar). We leave both to future work.



%
%

\clearpage

\bibliographystyle{splncs04}
\bibliography{main}

\clearpage
\appendix
\section{Additional Implementation Details}
For pretraining, we use two H200 GPUs in Distributed Data Parallel (DDP) mode, and for downstream fine-tuning, we use a single NVIDIA H200 GPU. We summarize all hyperparameter configurations in Tab.~\ref{tab:hyper}. For spatial clustering, $k_{\mathrm{nn}}$ and $\tau_d$ denote the neighbor count and maximum search radius used to construct a graph over PIT-normalized predicted centroids. A standard BFS algorithm decomposes the graph into connected components; components smaller than $\tau_{\mathrm{min}}$ are discarded to remove noisy groupings, and each remaining component is treated as a pseudo-instance.

\begin{table}[h]
\centering
\caption{Pretraining settings of \coolname{}}
\begin{tabular}{l |c }
\toprule
\textbf{Config} & \textbf{Value} \\
\midrule
Optimizer & AdamW  \\
Scheduler & Cosine annealing\\
Learning rate & 2e-4  \\
Weight decay & 4e-2  \\
Batch size & 16   \\
Mask Ratio & 0.6 \\
Mask Size & 1 m  \\
Warmup ratio & 0.05  \\
Teacher temperature & 0.035\\
Student temperature & 0.05\\
Training epochs & 50  \\
$\alpha_{zipf}$ & 1.3  \\
Warmup ratio of $\mathcal{L}_{\text{geo}}$ & 0.1  \\
Warmup ratio of activating predictor $G$ & 0.1\\
$\lambda$ & 0.1 \\
$ks$ (Occupancy Dilation) & 5  \\
$k$ (Token Initialization) & 3 \\
$k_{nn}$ (Spatial Clustering) & 20 \\
$\tau_d$ (Spatial Clustering)& 5 m \\
$\tau_{min}$ (Spatial Clustering) & 10\\
 Offset Direction Distribution (PIT-normalization) & Uniform(0, 1)  \\
 Offset Magnitude Distribution (PIT-normalization)& LogNormal($\mu$=1.4, $\sigma$=0.8) \\
 \bottomrule
\end{tabular}

\label{tab:hyper}
\end{table}

\section{Additional Experiments}
\subsection{Runtime Analysis}
Table~\ref{tab:runtime} reports total pretraining time for \coolname{} compared to DOS~\cite{abdelsamad2026dos} and PointINS~\cite{yang2026pointins}, measured on the same hardware under identical training configurations. \coolname{} requires 26 hours, compared to 20 hours for PointINS and 15 hours for DOS. The additional overhead relative to PointINS stems from the Neighborhood Sampling module and the ghost predictor $G$, which is incurred only during pretraining while downstream fine-tuning cost is identical to prior methods as both components are discarded. We consider this 30\% overhead over the strongest prior method acceptable given the consistent improvements in downstream 3D detection reported in the main results.
\begin{figure}[h]
\centering
\vspace{-4mm}
 \begin{minipage}{0.32\textwidth}
\centering
\captionof{table}{\textbf{Runtime Analysis:} we use the same hardware configuration across all models for fair comparison.}
\begin{tabular}{l|c }
\toprule
\textbf{Method} &\textbf{Time}\\
\midrule
DOS~\cite{abdelsamad2026dos} &  15 hours  \\
PointINS~\cite{yang2026pointins} &  20 hours\\
\coolname{} & 26 hours  \\

 \bottomrule
\end{tabular}
\label{tab:runtime}
\end{minipage}
\hfill
\begin{minipage}{0.65\textwidth}
\centering
\captionof{table}{\textbf{Effect of warmup ratio}: we tested different set of warmup ratios for two stages. The performance remains consistent.}
\begin{tabular}{l|c|c }
\toprule
\makecell[l]{\textbf{Warmup Ratio} \\ \textbf{(first stage, second stage)}} & \textbf{mAP} & \textbf{NDS}\\
\midrule
(0.1, 0.1) &  59.5 & 64.2  \\
(0.2, 0.2) &  59.3 & 64.1  \\
(0.1, 0.2) & 59.4 & 64.2  \\
(0.2, 0.1) & 59.2 & 64.0  \\

 \bottomrule
\end{tabular}
\label{tab:warmup_ratio}
\end{minipage}
\vspace{-4mm}
\end{figure}

\subsection{Effect of Warmup Ratio}
Table~\ref{tab:warmup_ratio} reports the sensitivity of \coolname{} to the two-stage warmup ratios. Performance remains stable across all tested 
configurations, with variations of less than 0.3 mAP and 0.2 NDS. The 
default configuration of (0.1, 0.1) achieves the best performance, and 
reducing or increasing either stage ratio has only marginal effect. These 
results confirm that \coolname{} is robust to the choice of warmup 
schedule as long as the two-stage order is preserved.

\subsection{Effect of Subsample Ratio}
Table~\ref{tab:subsample_ratio} reports the effect of the subsample ratio of unobserved voxels. Performance peaks at 0.5 and degrades slightly at both extremes, revealing a coverage-quality trade-off. At low ratios, newly sampled voxels are concentrated near instance boundaries where dilation first activates, providing reliable but spatially limited supervision to cover the unobserved object extent sufficiently. At high ratios, the number of sampled voxels increases significantly in the unobserved regions, where the distillation targets become noisier. As default, we set the ratio at 0.5 to balance these two effects.

\begin{figure}[h]
\centering

\begin{minipage}{0.4\textwidth}
\centering
\captionof{table}{\textbf{Effect of subsample ratio}: we test the different subsample ratio for sampling unobserved voxels in neighborhood sampling module.}
\begin{tabular}{l|c|c }
\toprule
\textbf{Subsample Ratio} & \textbf{mAP}&\textbf{NDS}\\
\midrule
0.2 &  58.9 & 63.9  \\
0.5 &  59.5 & 64.2  \\
0.8 & 59.2 & 64.1  \\

 \bottomrule
\end{tabular}
\label{tab:subsample_ratio}
\end{minipage}
\hfill
\begin{minipage}{0.57\textwidth}
\centering
\captionof{table}{\textbf{OOD robustness on nuScenes-C:} comparison of SSL methods under real-world corruptions. We use the models trained under decoder probing protocol and regard CenterPoint~\cite{yin2021center} as the unified baseline for computing metrics.}
\begin{tabular}{l|c|c }
\toprule
\textbf{Method} & \textbf{mRR} $\uparrow$&\textbf{mCE} 	$\downarrow$\\
\midrule
SONATA~\cite{wu2025sonata} &  96.1 &  88.6   \\
DOS~\cite{abdelsamad2026dos}  &  96.3 & 71.2  \\
PointINS~\cite{yang2026pointins} & 96.7 & 70.4\\
\coolname{} & \textbf{97.5} & \textbf{68.8} \\

 \bottomrule
\end{tabular}
\label{tab:ood}
\end{minipage}
\vspace{-4mm}
\end{figure}

\subsection{Out-of-Distribution (OOD) Generalization}
Table~\ref{tab:ood} evaluates robustness to out-of-distribution (OOD) conditions on the Robo3D benchmark~\cite{kong2023robo3d}, which measures 
resilience to real-world corruptions such as weather effects, sensor noise, and point cloud degradation. We report mean Resilience Rate (mRR) and mean Corruption Error (mCE), where higher mRR and lower mCE indicate 
greater robustness. \coolname{} achieves the best performance on both metrics, with 97.5 mRR and 68.8 mCE, outperforming all prior SSL methods by a substantial margin on mCE. This result suggests that explicitly 
modeling unobserved object geometry during pretraining not only improves detection under clean conditions but also produces representations that are more resilient to input corruptions.

\subsection{Effect of Kernel Size $ks$ and $k$ for Token Initialization}
\begin{figure}[b]
\vspace{-4mm}
    \centering
    \begin{subfigure}{0.48\textwidth}
        \centering
        \includegraphics[width=\linewidth]{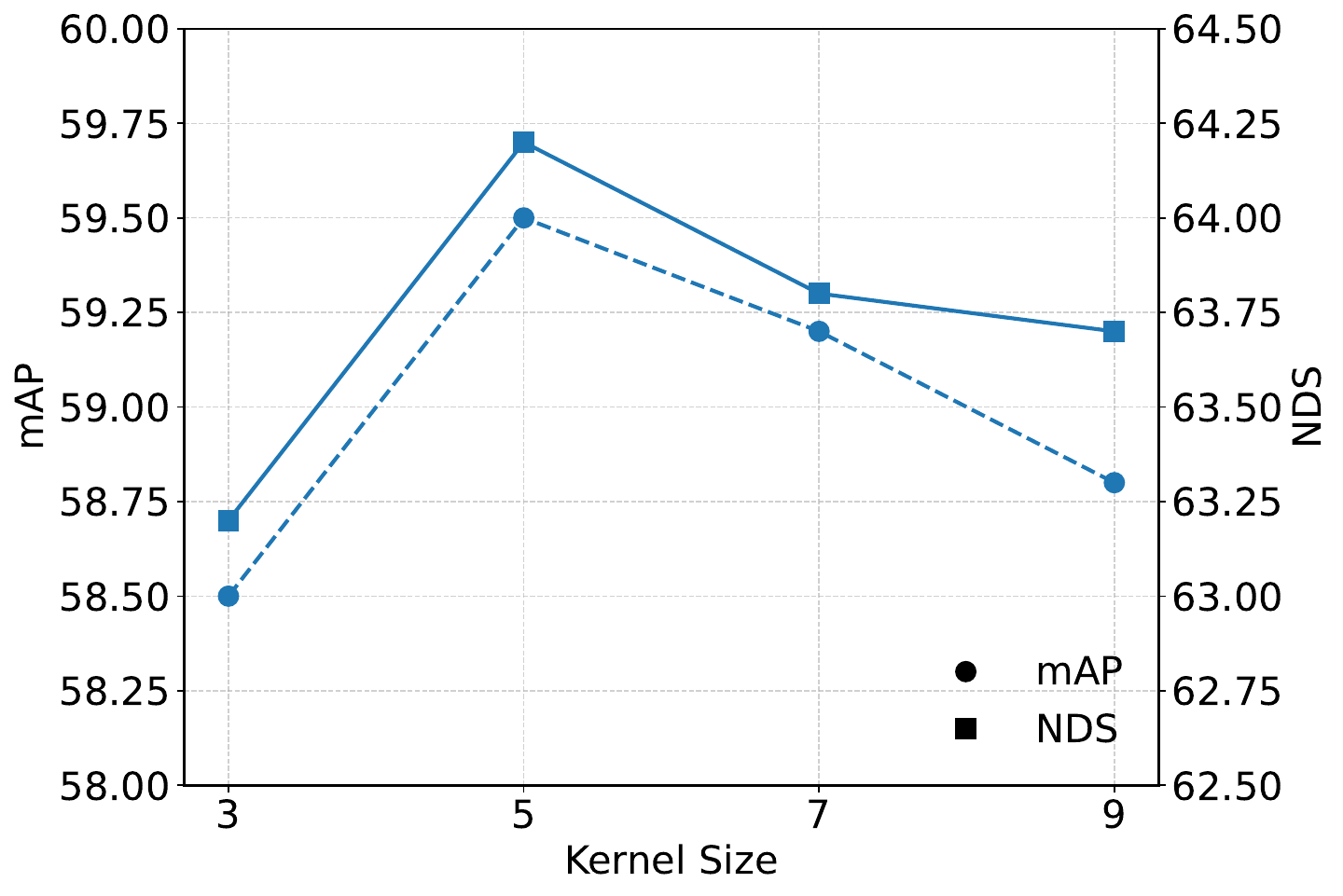}
        \caption{Occupancy Dilation ($ks$)}
        \label{subfig:kernel_size}
    \end{subfigure}
    \hfill
    \begin{subfigure}{0.48\textwidth}
        \centering
        \includegraphics[width=\linewidth]{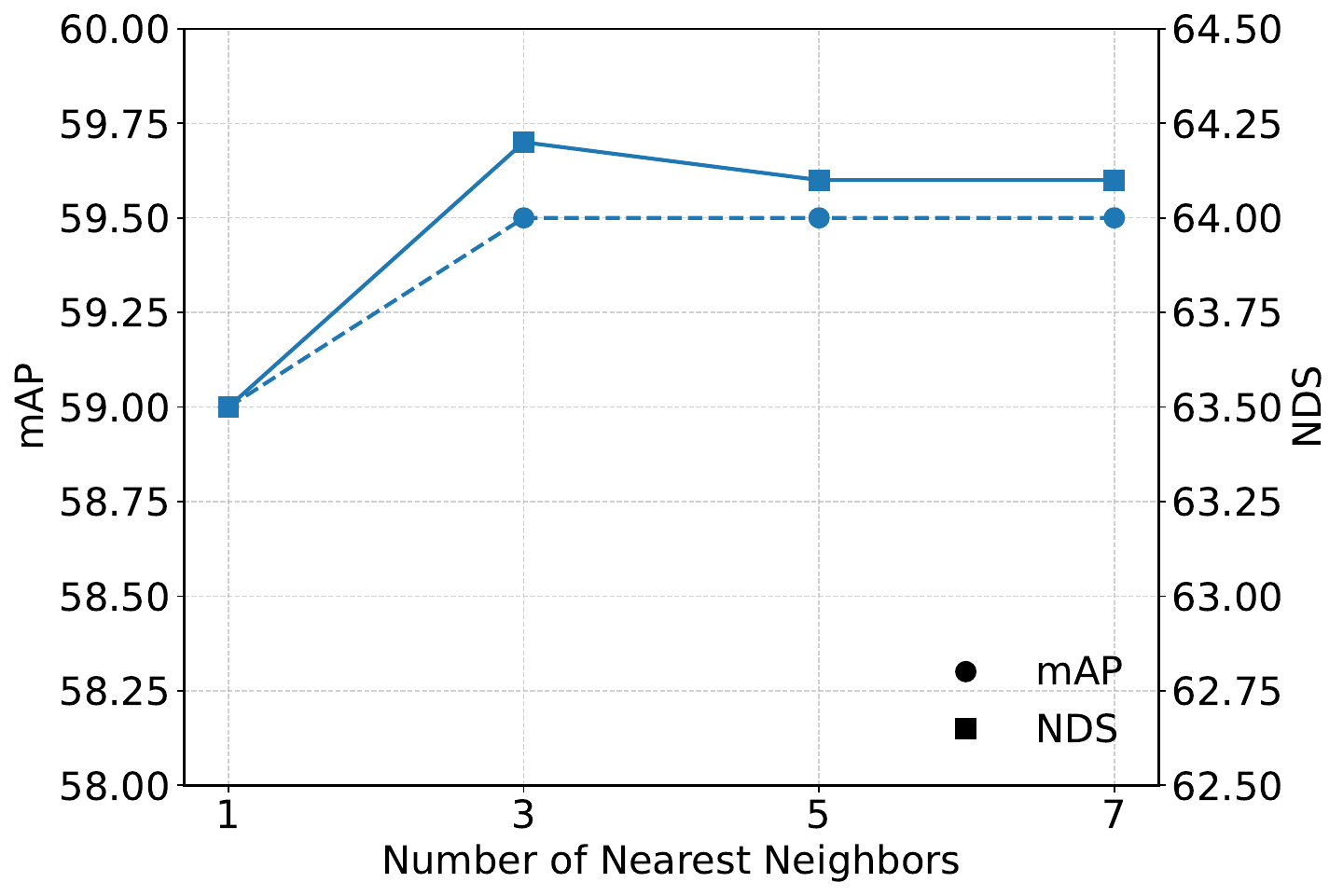}
        \caption{Token Initialization ($k$)}
        \label{subfig:k_token_initialization}
    \end{subfigure}

   \caption{Sensitivity analysis on two key hyperparameters: the dilation kernel size $ks$, which controls the spatial extent of occupancy dilation  around pseudo-instance proposals, and the neighbor count $k$, which determines how many observed encoder features are aggregated via KNN interpolation to initialize tokens at unobserved voxels.}
    \label{fig:sensitivity}
\end{figure}

Figure~\ref{fig:sensitivity} reports the sensitivity of \coolname{} to 
two key hyperparameters. For the dilation kernel size $ks$, performance 
peaks at $ks=5$ and degrades gradually for larger values, suggesting that 
moderate dilation captures sufficient unobserved object structure without 
introducing excessive background voxels that dilute the instance signal. 
For the neighbor count $k$, performance improves from $k=1$ to $k=3$ and 
remains stable beyond that, indicating that aggregating features from a 
small local neighborhood provides sufficient context for token 
initialization and that additional neighbors offer diminishing returns. 
Based on these results, we adopt $ks=5$ and $k=3$ as default 
configurations, both of which correspond to the peak or plateau of their 
respective sensitivity curves.

\end{document}